\documentclass[10pt,sigconf]{acmart}

\usepackage{booktabs} 
\usepackage{multirow}
\usepackage{multicol}
\usepackage{booktabs}
\usepackage[utf8]{inputenc}
\usepackage{graphicx}
\usepackage{subfigure}
\usepackage{fancyhdr}

\graphicspath{{figure/}{figures/}}

\setcopyright{acmlicensed}
\acmDOI{}
\acmISBN{}
\acmConference[DAPA '19]{DAPA 2019 WSDM Workshop on Deep matching in Practical Applications}{February 15th, 2019}{Melbourne, Australia}
\acmYear{2019}
\copyrightyear{2019}

\acmPrice{}

\begin{document}
\title{Attention Boosted Sequential Inference Model}

\author{Guanyu Li}
\orcid{1234-5678-9012}
\affiliation{%
  \institution{School of Computer and Information Technology \& Beijing Key Lab of Traffic Data Analysis and Mining, Beijing Jiaotong University}
  \streetaddress{}
  \city{Beijing}
  \state{China}
  \postcode{100044}
}
\email{17120379@bjtu.edu.cn}

\author{Pengfei Zhang}
\orcid{1234-5678-9012}
\affiliation{%
  \institution{School of Computer and Information Technology \& Beijing Key Lab of Traffic Data Analysis and Mining, Beijing Jiaotong University}
  \streetaddress{}
  \city{Beijing}
  \state{China}
  \postcode{100044}
}
\email{18120448@bjtu.edu.cn}

\author{Caiyan Jia*}
\orcid{1234-5678-9012}
\affiliation{%
  \institution{School of Computer and Information Technology \& Beijing Key Lab of Traffic Data Analysis and Mining, Beijing Jiaotong University}
  \streetaddress{}
  \city{Beijing}
  \state{China}
  \postcode{100044}
}
\email{cyjia@bjtu.edu.cn}

\renewcommand{\shortauthors}{Guanyu Li et al.}

\begin{abstract}
Attention mechanism has been proven effective on natural language processing.  This paper proposes an attention boosted natural language inference model named aESIM by adding word attention and adaptive direction-oriented attention mechanisms to the traditional Bi-LSTM layer of natural language inference models, e.g. ESIM. This makes the inference model aESIM has the ability to effectively learn the representation of words and model the local subsentential inference between pairs of premise and hypothesis. The empirical studies on the SNLI, MultiNLI and Quora benchmarks manifest that aESIM is superior to the original ESIM model.
\end{abstract}

%
%



\keywords{natural language processing, deep learning, natural language inference, Bi-LSTM}

\maketitle

\section{Introduction}

Natural language inference (NLI) is an important and significant task in natural language processing (NLP). It concerns whether a hypothesis can be inferred from a premise, requiring understanding of the semantic similarity between the hypothesis and the premise to discriminate their relation \cite{lan2018toolkit}. Table \ref{tab:samples} shows several samples of natural language inference from SNLI (Stanford Natural Language Inference) corpus \cite{bowman2015large}.

 In the literature, the task of NLI is usually viewed as a relation classification. It learns the relation between a premise and a hypothesis in a large training set, then predicts the relation between a new pair of premise and hypothesis. The existing methods of NLI can be roughly partitioned into two categories: feature-based models \cite{bowman2015large} and neural network-based models \cite{yang2016hierarchical,chen2016enhanced}. Feature-based models represent a premise and a hypothesis by their unlexicalized and lexicalized features, such as $n$-gram length and the real-valued feature of length difference, then train a classifier to perform relation classification. Recently, end-to-end neural network-based models have drawn worldwide attention since they have demonstrated excellent performance on quite a few NLP tasks including machine translation, natural language inference, etc.

 \begin{table}[h]
\begin{tabular}{|p{2.5cm}|p{3cm}|l|}
\hline
premise                     & hypothesis                        & relationship  \\ \hline
Wet brown dog swims towards & A dog is playing fetch in a pond. & neutral       \\ \cline{2-3}
camera.                     & A dog is in the water.            & entailment    \\ \cline{2-3}
                            & The dog is sleeping in his bed.   & contradiction \\ \hline
\end{tabular}
 \caption{Samples from the SNLI corpus}
 \label{tab:samples}
\end{table}

\begin{figure}[h]
    \centering
    \subfigure[sentence encoding model]{
    \begin{minipage}[b]{0.5\textwidth}
    \centering
        \includegraphics[width=0.8\textwidth,height=0.5\textwidth]{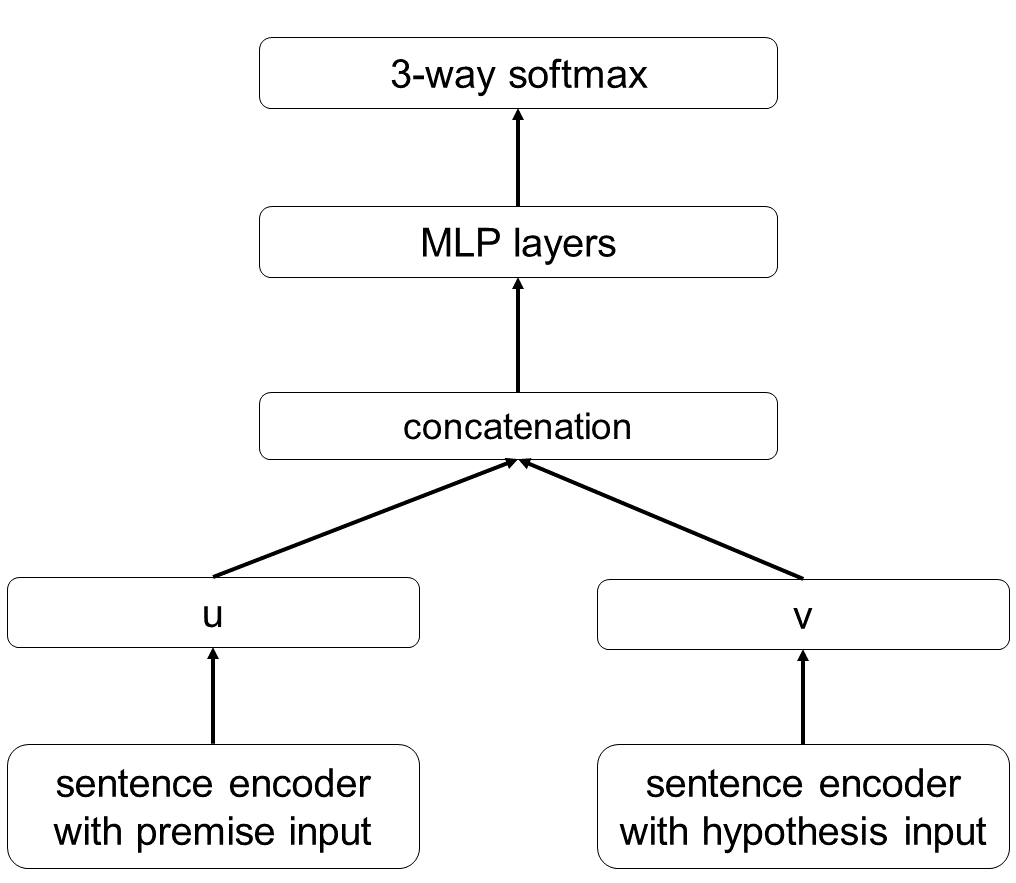}
    \end{minipage}
    }
    \subfigure[sentence interaction-aggregation model]{
    \begin{minipage}[b]{0.5\textwidth}
    \centering
        \includegraphics[width=0.8\textwidth,height=0.5\textwidth]{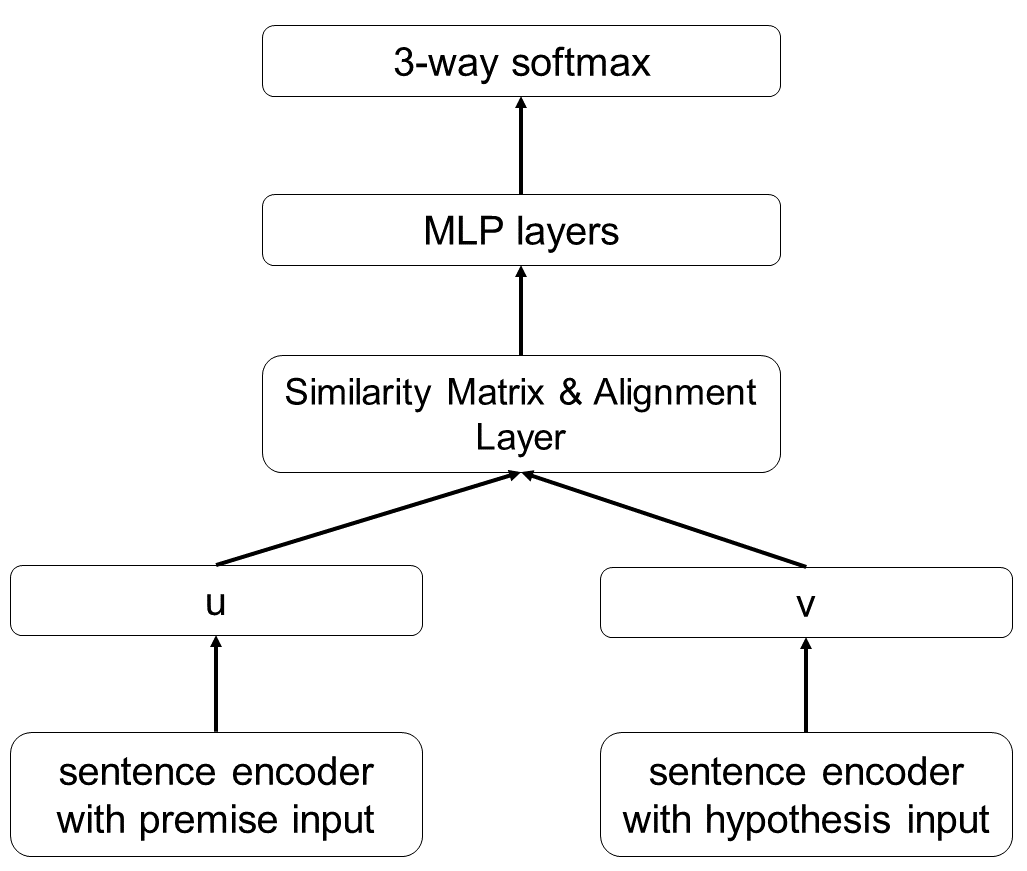}
    \end{minipage}
    }

    \caption{Two types of neural network-based models}
    \label{fig:different type model}
\end{figure}

 On the basis of their model structures, we can divide neural network-based models for NLI into two classes \cite{lan2018toolkit}, sentence encoding models and sentence interaction-aggregation models. The architectures of the two types of models are shown in Figure \ref{fig:different type model}.

Sentence encoding models \cite{conneau2017supervised,talman2018natural,im2017distance,shen2018reinforced} (their main architecture is shown in Figure \ref{fig:different type model}.a) independently encode a pair of sentences, a premise and a hypothesis using pre-trained word embedding vectors, then learn semantic relation between two sentences with a multi-layer perceptron (MLP). In these models, LSTM (Long Short-Term Memory networks) \cite{cheng2016long}, its variants GRU (Gated Recurrent Units) \cite{chung2014empirical} and Bi-LSTM, are usually utilized to encode the sentences since they were capable of learning long-term dependencies inside sentences. For example, Conneau {\it et al.} proposed a generic NLI training scheme and compared several sentence encoding architectures: LSTM or GRU, Bi-LSTM with mean/max pooling, self-attention network and hierarchical convolutional networks \cite{conneau2017supervised}. The experimental results demonstrated that the Bi-LSTM with max pooling achieved the best performance. Talman {\it et al.} designed a hierarchical Bi-LSTM max pooling (HBMP) model to encode sentences \cite{talman2018natural}. This model applied parameters of one Bi-LSTM to initialize the next Bi-LSTM to convey information, which shown better results than the model with a single Bi-LSTM. Besides LSTM, attention mechanisms could also be used to boost the effectiveness of sentence encoding. The model developed by Ghaeini {\it et al.} added self-attention to LSTM model, and achieved better performance \cite{ghaeini2018dr}.

Sentence interaction-aggregation models  \cite{parikh2016decomposable,wang2017bilateral,kim2018semantic,lan2018toolkit} (their main architecture is shown in Figure \ref{fig:different type model}.b) learn vector representations of pairs of sentences in the way similar to sentence encoding models and calculate pairwise word interaction matrix between two sentences using the newly updated word vectors, and then the matching results are aggregated into a vector to make the final decision. Compared with sentence encoding model, sentence interaction-aggregation models aggregate word similarities between a pair of sentences, are capable of capturing the relevant information between two sentences, a premise and a hypothesis. Bahdanau {\it et al.}  translated and aligned text simultaneously in machine translation task  \cite{bahdanau2014neural}, innovatively introducing attention mechanism to natural language process (NLP). He {\it et al.} designed a pairwise word interaction model (PWIM)  \cite{he2016pairwise}, which made full use of word-level fine-grained information. Wang {\it et al.} put forward a bilateral multi-perspective matching (BiMPM) model \cite{wang2017bilateral}, focusing on various matching strategies that could be seen as different types of attention. The empirical studies of Lan {\it et al.} \cite{lan2018toolkit} and Chen {\it et al.} \cite{chen2016enhanced} concluded that sentence interation-aggregation models, especially ESIM (Enhanced Sequential Inference Model), a carefully designed sequential inference model based on chain LSTMs, outperformed all previous sentence encoding models.

Although ESIM has achieved excellent achievements, this model doesn’t consider the attention along the words in a sentence in its Bi-LSTM layer. Word attention can characterize the different contribution of each word. Therefore, it will be beneficial to put word attention into the Bi-LTSM layer. Moreover, the orientation of the words represents the direction of the information flow, either forward or backward, should not be ignored. In traditional Bi-LSTM model, the forward and the backward vectors learnt by Bi-LSTM are simply jointed. It's necessary to consider whether each orientation (forward or backward) has different importance on word encoding, thus adaptively joint the two orientation vectors together with different weights. Therefore, in this study, using ESIM model as the baseline, we add an attention layer behind each Bi-LSTM layer, then use an adaptive orientation embedding layer to jointly represent the forward and backward vectors. We name this attention boosted Bi-LSTM as Bi-aLSTM, and denote the modified ESIM as aESIM. Experimental results on SNLI, MultiNLI \cite{williams2017broad} and  Quora \cite{wang2017bilateral} benchmarks have demonstrated better performance of aESIM model than that of the baseline ESIM and the other state-of-the-art models. We believe that the architecture of Bi-aLSTM has potentially to be used in other NLP tasks such as text classification, machine translation and so on.

This paper is organized as follows. We introduce the general frameworks of ESIM and aESIM in Section 2. We describe the datasets and the experiment settings, and analyze our experimental results in Section 3. We then draw conclusions in Section 4.

\section{Attention Boosted Sequential Inference Model}
Supposed that we have two sentences $p=({{p}_{1}},\cdots ,{{p}_{{{l}_{p}}}})$ and $q=({{q}_{1}},\cdots ,{{q}_{{{l}_{q}}}})$,
where $p$ represents premise and $q$ represents hypothesis. The goal is to predict the label $y$ meaning for their relation.
\subsection{ESIM model}
Enhanced Sequential Inference Model (ESIM) \cite{cheng2016long} is composed of four main components: input encoding layer, local inference modeling layer, inference composition layer and classification layer.

In the input encoding layer, ESIM first uses Bi-LSTM layer to encode input sentence pairs (Equations 1-2), which can be initialized using pre-trained word embeddings (e.g. Glove 840B vectors \cite{pennington2014glove}), where $(p, i)$ is the word embedding vector of the $i$-th word in $p$, $(q,i)$ is that of word in $q$.

\begin{equation}
 \overline{{{p}_{i}}}=Bi\text{-}LSTM(p,i),\forall i\in [1,\cdots ,{{l}_{p}}]\
\end{equation}
\begin{equation}
     \overline{{{q}_{j}}}=Bi\text{-}LSTM(q,j),\forall j\in [1,\cdots ,{{l}_{q}}]
\end{equation}

Secondly, ESIM implements the local inference layer for enhancing the sentence information. First it calculates a similarity matrix $M$ based on $\overline{p}$ and $\overline{q}$.
\begin{equation}
M={{\overline{p}}^{T}}\overline{q}
\end{equation}
It then gets the new expression for $p$ and $q$ with the equation below:
\begin{equation}
\widetilde{{{p}_{i}}}=\sum\limits_{j=1}^{{{l}_{q}}}{\frac{\exp ({{M}_{ij}})}{\sum\limits_{k=1}^{{{l}_{q}}}{\exp ({{M}_{ik}})}}}\overline{{{q}_{j}}},\forall i\in [1,\cdots ,{{l}_{p}}]
\end{equation}
\begin{equation}
 \widetilde{{{q}_{j}}}=\sum\limits_{i=1}^{{{l}_{p}}}{\frac{\exp ({{M}_{ij}})}{\sum\nolimits_{k=1}^{{{l}_{p}}}{\exp ({{M}_{kj}})}}}\overline{{{p}_{i}}},\forall j\in [1,\cdots ,{{l}_{q}}]
\end{equation}
where $\widetilde{p}$ and  $\widetilde{q}$ represent the weighted summation of   $\overline{p}$ and $\overline{q}$. It further enhances the local inference information collected as below.
\begin{equation}
    {{m}_{p}}=[\overline{p};\widetilde{p};\overline{p}-\widetilde{p};\overline{p}\odot \widetilde{p}]
\end{equation}
\begin{equation}
    {{m}_{q}}=[\overline{q};\widetilde{q};\overline{q}-\widetilde{q};\overline{q}\odot \widetilde{q}]
\end{equation}

After the enhancement of local inference, another Bi-LSTM layer is used to capture local inference information and their context for inference composition.

Instead of summation adopted by Parikh {\it et al.} \cite{parikh2016decomposable}, ESIM proposes to compute both max and average pooling and feeds the concatenate fixed length vector to the final classifier: a fully connected multi-layer perceptron.

\begin{figure}[h]
    \centering
    \includegraphics[width=\linewidth]{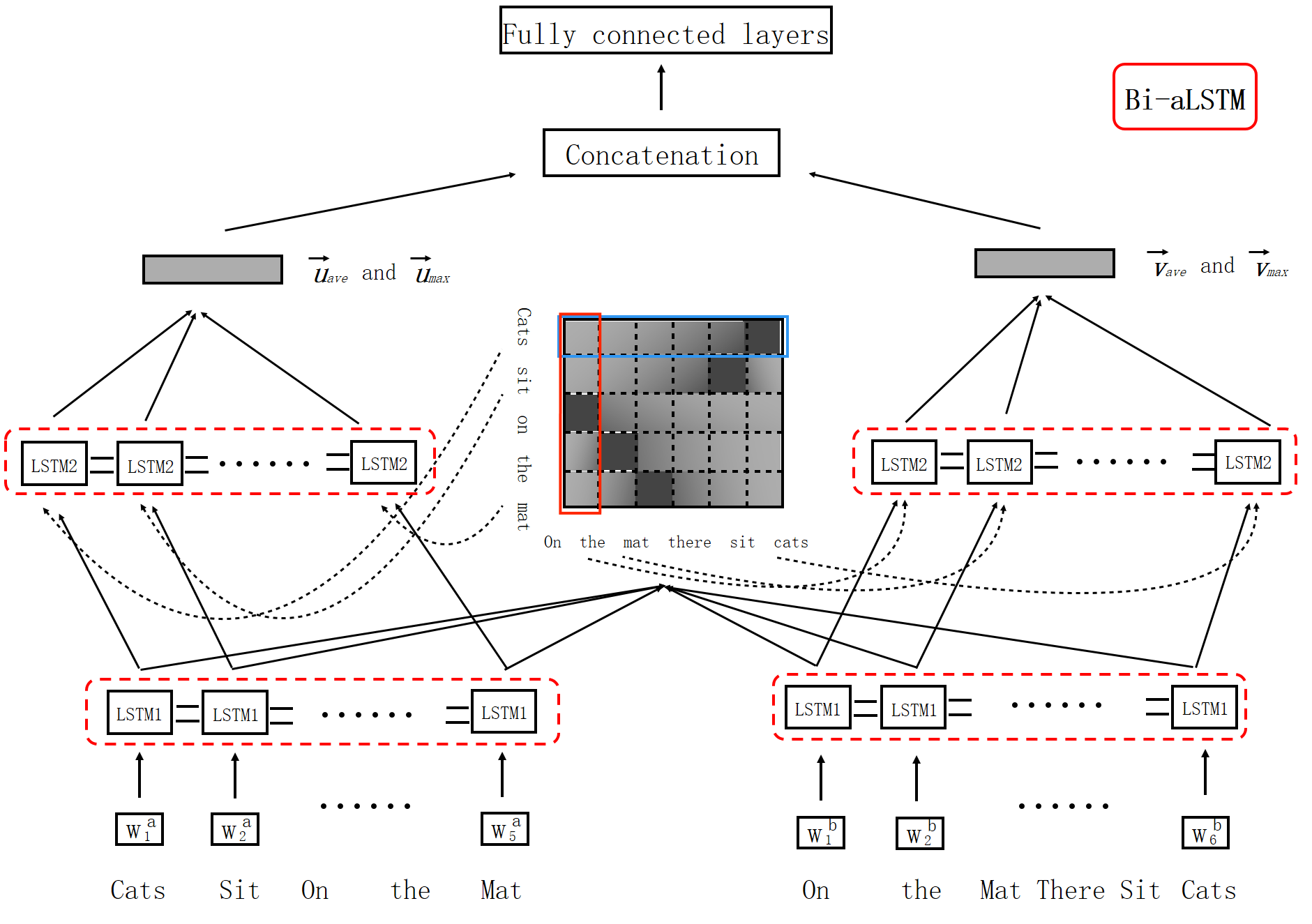}
    \caption{ESIM and aESIM model architectures}
    \label{fig:ESIM-figl}
\end{figure}

Figure \ref{fig:ESIM-figl} shows a high-level view of the ESIM architecture, where the bottom LSTM1 layer of Figure \ref{fig:ESIM-figl} is the input encoding layer, the middle part with LSTM2 layer is the local inference layer, the upper part is the inference composition layer.

\subsection{aESIM model}

\begin{figure}[t]
    \centering
    \includegraphics[width=\linewidth]{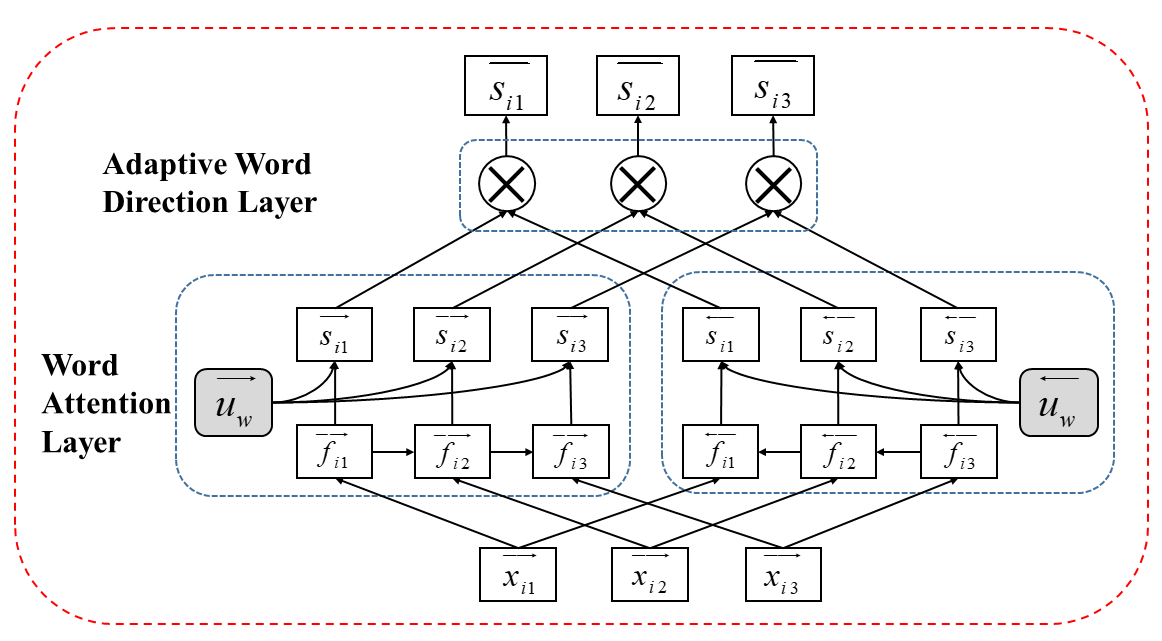}
    \caption{The structure of Bi-aLSTM including input layer, word attention layer and adaptive word direction layer.}
    \label{fig:ALSTM-figl}
\end{figure}

The overall architecture of our newly proposed attention boosted sequential inference model (named aESIM) based on ESIM is similar to ESIM. In detail, aESIM also consists of four main parts: encoding layer, local inference modeling layer, decoding layer and classification layer. The only difference between ESIM and aESIM is that we substitute the two Bi-LSTM layers (LSTM1 and LSTM2) in ESIM with two Bi-aLSTM layers in aESIM. Therefore, as illustrated in Figure \ref{fig:ESIM-figl}, the layers with red-dotted circles in ESIM will be replaced by the Bi-aLSTM layers shown in the right upper corner of the Figure \ref{fig:ESIM-figl} and the details of Bi-aLSTM can be found in Figure \ref{fig:ALSTM-figl}.

Given the word vector ${{x}_{il}},l\in [1,T]$ of the $l$-th word in sentence $i$, which can be obtained by pre-trained word embeddings such as Glove 840B vectors \cite{pennington2014glove} in the first Bi-aLSTM layer or obtained from the local inference modeling layer in the second Bi-aLSTM layer. We utilize a forward LSTM layer and a backward LSTM layer to collect both direction information $\overrightarrow{f}$ and $\overleftarrow{f}$ .

\begin{equation}
{{\overrightarrow{f}}_{il}}=\overrightarrow{LSTM}({{x}_{il}}),l\in \left[ 0,T \right]
\end{equation}
\begin{equation}
{{\overleftarrow{f}}_{il}}=\overleftarrow{LSTM}({{x}_{il}}),l\in \left[ 0,T \right]
\end{equation}
As described in introduction section, in the following newly proposed Bi-aLSTM, we add word attention and additive operation on both orientations of traditional Bi-LSTM layer.

\textbf{Word attention layer}

It’s obvious that not all words contribute equally to the representation of a sentence. Attention mechanism, which is introduced in \cite{yang2016hierarchical}, is extremely effective to extract vital words from the whole sentence, and is particularly beneficial to generate the sentence vector. Therefore, we use the following attention mechanism after we get $\overrightarrow{f}$ and $\overleftarrow{f}$.

Suppose $f_{il}\in \{\overrightarrow{f}_{il}, \overleftarrow{f}_{il}\}$, we then have
\begin{equation}
{{u}_{il}}=\tanh (W{{f}_{il}}+b)
\end{equation}
\begin{equation}
{{\alpha }_{il}}=\frac{\exp (u_{il}^{T}{{u}_{w}})}{\sum\limits_{l}{\exp (u_{il}^{T}{{u}_{w}})}}
\end{equation}
\begin{equation}
{{s}_{il}}={{\alpha }_{il}}*{{f}_{il}}
\end{equation}
where ${{u}_{il}}$ is obtained after one-layer MLP for the input $f_{il}$, ${{\alpha }_{il}}$ is the importance of word $l$, is calculated by the SoftMax unit on the context vector ${{u}_{w}}$ of the sentence $i$ which is randomly initialized and modified during the training, ${{s}_{il}}$ is the attention enhanced vector through multiplying the weight ${{\alpha }_{il}}$ and original vector ${{f}_{il}}$, where ${s}_{il}\in \{\overrightarrow{{s}_{il}}, \overleftarrow{{s}_{il}}\}$ correspond to the forward vector $\overrightarrow{f}_{il}$ and the backward vector $\overleftarrow{f}_{il}$, respectively.

\textbf{Adaptive word direction layer}

In traditional Bi-LSTM model, the forward and the backward vectors of a word are considered to have equal importance on the word representation. The model simply connects the forward and backward vectors head and tail without weighing their importance. For a word in different direction or orientation, the former and the latter words are reversed. Thus, different direction vectors of a word make different contribution to the representation, especially the words in a long sentence. Therefore, we propose a new adaptive direction layer to learn the contribution of different directions for a single word.

Formally, given two direction word vectors $\overrightarrow{{{s}_{il}}}$ and $\overleftarrow{{{s}_{il}}}$, the whole word vector can be expressed as:
\begin{equation}
\overline{{s}_{il}}=g[({{W}_{F}}*\overrightarrow{{{s}_{il}}}+{{b}_{F}})\centerdot ({{W}_{B}}*\overleftarrow{{{s}_{il}}}+{{b}_{B}})]
\end{equation}
where,${{W}_{*}}$ and ${{b}_{*}}$ denote weight matrix and the bias, $g$ denotes the nonlinear function, $[\centerdot ]$ denotes the concentration. All the parameters can be learned during training. Then we can get the whole sentence vector as below:
\begin{equation}
 \overline{{{p}_{i}}}=Bi\text{-}aLSTM(\overline{{s}_{il}}),\forall i\in [1,\cdots ,{{l}_{p}}]\
\end{equation}
\begin{equation}
     \overline{{{q}_{j}}}=Bi\text{-}aLSTM(\overline{{s}_{jl}}),\forall j\in [1,\cdots ,{{l}_{q}}]
\end{equation}
This word and orientation enhanced Bi-LSTM is called Bi-aLSTM. Its whole architecture is shown in the Figure \ref{fig:ALSTM-figl}, is applied in ESIM model to replace the two Bi-LSTM layers for the task of natural language inference. Besides, this Bi-aLSTM can be used to other natural language processing tasks and our preliminary experiments have demonstrated that Bi-aLSTM is capable of improving the performance of Bi-LSTM models on sentimental classification task (for space limitation, this results will not be shown in the paper).





\section{Experiment Setup}
\subsection{Datasets}
We evaluated our model on three datasets: the Stanford Natural Language Inference (SNLI) corpus, the Multi-Genre Natural Language Inference (MultiNLI) corpus, and Quora duplicate question dataset. We selected these three relatively large corpora out of eight corpora in \cite{lan2018toolkit} since deep learning models usually show better generalization ability on large training sets and produce more convincing results than on small training sets.

\textbf{SNLI} The Stanford Natural Language Inference (SNLI) corpus contains 570,152 sentence pairs, including 549K training pairs, 10K validation pairs and 10K testing pairs. Each pair has one of relation classes (entailment, neutral, contradiction and ‘-’). The ‘-’ class indicates there is no conclusion between the two sentences. Consequently, we remove all pairs with relation '-' during training, validating and testing processes.

\textbf{MultiNLI} This corpus is a crowd-sourced collection of 433K sentence pairs annotated with textual entailment information. The corpus is modeled on the SNLI corpus, but differs in that covers a range of genres of spoken and written text, and supports a distinctive cross-genre generation evaluation.

\textbf{Quora} The Quora dataset contains 400,000 question pairs. The task of this corpus is to judge whether the two sentences means the same affair.

\subsection{Setting}

We use the validation set to select models for testing. The hyper-parameters of aESIM model are listed as follows. We use the Adam method \cite{kingma2014adam} for optimization. The first momentum is set to be 0.9 and the second 0.999. The initial learning rate is set to 0.0005, and the batch size is 128. The dimensions of all hidden states of Bi-aLSTM and word embedding are 300. We employ non-linearity function $f=selu$ \cite{klambauer2017self} replacing rectified linear unit $ReLU$ on account of its faster convergence rate. Dropout rate is set to 0.2 during training. We use pre-trained 300-D Glove 840B vectors \cite{pennington2014glove} to initialize word embeddings. Out-of-vocabulary (OOV) words are initialized randomly with Gaussian samples. All vectors are updated during training.

\begin{figure*}[h]
    \centering
    \subfigure[contradiction pair]{
    \begin{minipage}[t]{0.3\textwidth}
    \centering
        \includegraphics[width=\textwidth,height=5cm]{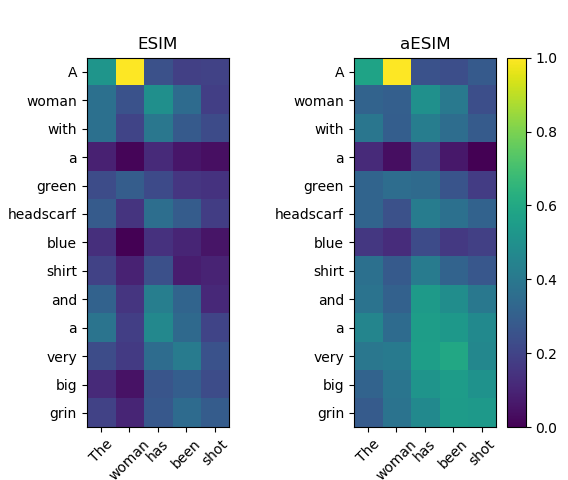}
    \end{minipage}
    }
    \subfigure[entailment pair]{
    \begin{minipage}[t]{0.3\textwidth}
    \centering
        \includegraphics[width=\textwidth,height=5cm]{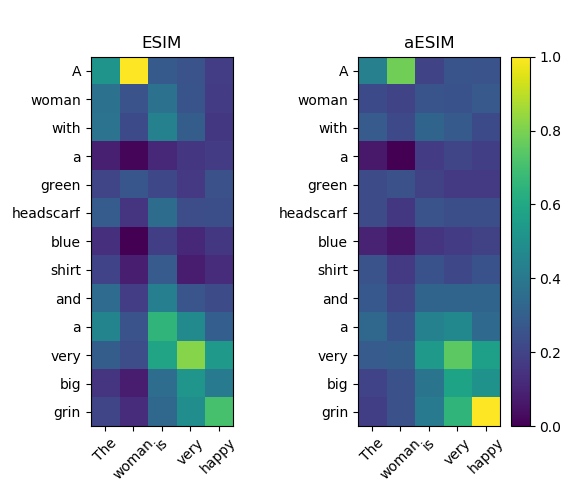}
    \end{minipage}
    }
    \subfigure[neutral pair]{
    \begin{minipage}[t]{0.3\textwidth}
    \centering
        \includegraphics[width=\textwidth,height=5cm]{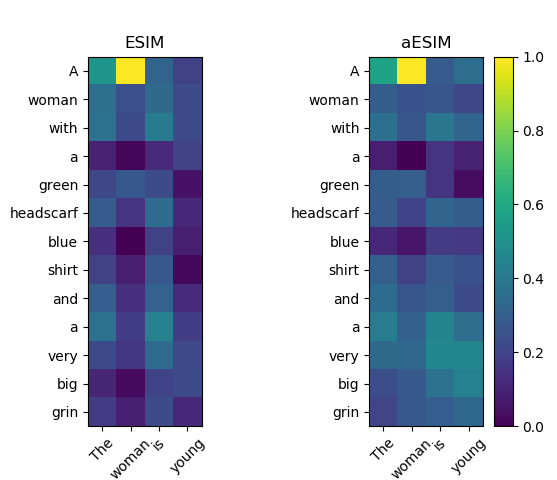}
    \end{minipage}
    }
    \caption{Attention visualization}
    \label{fig:attention}
\end{figure*}

\subsection{Experiment results}
Except for comparing our method aESIM with ESIM, we listed the experimental results of methods with their references in Table \ref{tab:snli_result} on SNIL. In Table \ref{tab:snli_result}, the method in the first block is a traditional feature engineering method, those in the second are the sentence vector-based models, those in the third are attention-based models, and ESIM and our aESIM are shown in the fourth block. Where the results of ESIM and aESIM are implemented by ourselves on Keras, the results of the others are taken from their original publications. We then compare the baseline models, CBOW, Bi-LSTM with ESIM and our aESIM on MultiNLI corpus shown In Table \ref{tab:results_multinli}, where the results of the baselines are taken from \cite{williams2017broad}. Finally,we compare several types of CNN and RNN models on Quroa corpus shown in Table \ref{tab:rs_quoral}, the results of theses CNN and RNN models are taken from \cite{wang2017bilateral}. The accuracy (ACC) of each method is measured by the commonly used precision score \footnote{https://nlp.stanford.edu/projects/snli/}, and the methods with the best accuracy are marked in bold.

\begin{table}[h]
    \centering
    \begin{tabular}{l|c}
        \hline
        \textbf{Models} & \textbf{Acc} \\
        \hline
         Unlexicalized + Unigram and bigram features \cite{bowman2015large} & 78.2 \\
        \hline
        300D LSTM encoders \cite{bowman2015large} & 80.6 \\
        300D NTI-SLSTM-LSTM encoders \cite{munkhdalai2017neural} & 83.4\\
        4096D Bi-LSTM with max-pooling \cite{conneau2017supervised} & 84.5 \\
        300D Gumbel TreeLSTM encoders \cite{choi2018learning} & 85.6\\
        512D Dynamic Meta-Embeddings \cite{kiela2018dynamic} & 86.7 \\
        \hline
        100D DF-LSTM17 \cite{liu2016deep}&84.6\\
        300D LSTMN with deep attention fusion \cite{cheng2016long}&85.7\\
        BiMPM \cite{wang2017bilateral}&87.5\\
        \hline
        ESIM & 87.3\\
        {\bf aESIM}& {\bf 88.1}\\
        \hline
    \end{tabular}
    \caption{The accuracy (\%) of the methods on SNLI}
    \label{tab:snli_result}
\end{table}

\begin{table}[h]
	\centering
	\setlength{\tabcolsep}{7.5mm}{
	
	\begin{tabular}{c|c|c}
		\hline
		\multirow{2}{*}{\textbf{Models}}    & \multicolumn{2}{c}{\textbf{Accuracy (\%)}}             \\
		\cline{2-3} & Matched       & Mismatched    \\
		\hline
		\multicolumn{1}{l|}{CBOW} & 64.8 & 64.5 \\
		\hline
		\multicolumn{1}{l|}{Bi-LSTM} & 66.9 & 66.9 \\
	    \hline
	    \multicolumn{1}{l|}{ESIM}  &73.4& 73.5 \\
	    \hline
	    \multicolumn{1}{l|}{\bf aESIM} &{\bf 73.9} & {\bf 73.9} \\
	    \hline
	\end{tabular}}
	\caption{ The accuracy (\%) of the methods on MultiNLI}
	\label{tab:results_multinli}
\end{table}

\begin{table}[h]
    \centering
    \setlength{\tabcolsep}{7.5mm}{
    \begin{tabular}{l|c}
        \hline
         \textbf{Models}& \textbf{Accuracy (\%)} \\
         \hline
         Siamese-CNN& 79.60 \\
         Multi-perspective-CNN & 81.38 \\
         Siamese-LSTM & 82.58 \\
         Multi-Perspective-LSTM & 83.21 \\
         L.D.C & 85.55 \\
         \hline
         ESIM & 86.98 \\
         {\bf aESIM} & {\bf 88.01} \\
         \hline
    \end{tabular}}
    \caption{ The accuracy (\%) of the methods on Quora}
    \label{tab:rs_quoral}
\end{table}

According to the results in Tables 2-4, aESIM model achieved 88.1\% on SNLI corpus, elevating 0.8 percent higher than ESIM model. It promoted almost 0.5 percent accuracy and outperformed the baselines on MultiNLI. It also achieved 88.01\% on Quora. Therefore, we concluded that aESIM with further word attention and word orientation operation was superior to ESIM model.

\subsection{Attention visualization}
We selected three types of sentence pairs from a premise and its three hypothesis sentences in the test set of SNLI corpus as shown in Figure \ref{fig:attention}, where the premise sentence is ‘A woman with a green headscarf, blue shirt and a very big grin’, and three hypothesis sentences are ‘the woman has been shot’, ‘the woman is very happy’ and ‘the woman is young’ with relation labels ‘contradiction’, ‘entailment’, and ‘neutral’, respectively. Each pair of sentences has their key word pairs: grin-shot, grin-happy and grin-young, which determines whether the premise can entail the hypothesis. Figures 4.a-4.c are the visualization of the attention layer between sentence pairs after the Bi-LSTM layer in ESIM model and that after Bi-aLSTM layer in aESIM model for contrasting ESIM and aESIM. By doing so, we could understand how the models judge the relation between two sentences.

In each Figure, the brighter the color, the higher the weight is. We could conclude that our aESIM model had the higher weight than ESIM model on each key word pair, especially in Figure \ref{fig:attention}.b, where the similarity of ‘happy’ and ‘grin’ in aESIM model is much higher than that in ESIM model. Therefore, our aESIM model was able to capture the most important word pair in each pair of sentences.

\section{Conclusion}
In this study, we propose an improved version of ESIM named aESIM for NLI. It modifies the Bi-LSTM layer to collect more information. We evaluate our aESIM model on three NLI corpora. Experimental results show that aESIM model achieves better performance than ESIM model. In the future, we will evaluate how attention mechanisms can be applied on other tasks and explore a way to use less time and space with guaranteed accuracy.

\section*{Acknowledgement}
This work is supported in part by the National Nature Science Foundation of China (No. 61876016 and No. 61632004), the Fundamental Research Funds for the Central Universities (No. 2018JBZ006).

\bibliographystyle{ieeetr}

\bibliographystyle{acm}
\bibliography{aesim}

\end{document}